\documentclass[10pt,twocolumn]{article}

\usepackage{iccv}
\usepackage{xcolor}
\usepackage{times}
\usepackage{epsfig}
\usepackage{enumitem}
\usepackage{graphicx}
\usepackage{amsmath}
\usepackage{amssymb}
\usepackage{bbm}
\usepackage{booktabs}
\usepackage{multirow}
\usepackage{pifont}
\newcommand{\cmark}{\ding{51}}%
\newcommand{\xmark}{\ding{55}}
\usepackage[pagebackref=true,breaklinks=true,letterpaper=true,colorlinks,bookmarks=false]{hyperref}

\iccvfinalcopy 

\def\iccvPaperID{9250} 

\ificcvfinal\fi

\begin{document}

\title{Contrastive Learning for Label Efficient Semantic Segmentation}
\author{Xiangyun Zhao${}^{1,2}$\thanks{This work was done when Xiangyun Zhao was interning at Google.}\\
        {\tt\small zhaoxiangyun915@gmail.com}
        \and
        Raviteja Vemulapalli${}^{2}$\\
        {\tt\small ravitejavemu@google.com}
        \and
        Philip Andrew Mansfield${}^{2}$\\
        {\tt\small memes@google.com}
        \and
        Boqing Gong${}^{2}$\\
        {\tt\small bgong@google.com }
        \and
        Bradley Green${}^{2}$\\
        {\tt\small brg@google.com }
        \and
        Lior Shapira${}^{2}$\\
        {\tt\small liorshap@google.com }
        \and
        Ying Wu${}^{1}$\\
        {\tt\small yingwu@northwestern.edu }
        \and
        ${}^{1}$Northwestern University \, ${}^{2}$Google Research
}

\maketitle
\ificcvfinal\thispagestyle{empty}\fi

\begin{abstract}
Collecting labeled data for the task of semantic segmentation is expensive and time-consuming, as it requires dense pixel-level annotations. While recent Convolutional Neural Network (CNN) based semantic segmentation approaches have achieved impressive results by using large amounts of labeled training data, their performance drops significantly as the amount of labeled data decreases. This happens because deep CNNs trained with the de facto cross-entropy loss can easily overfit to small amounts of labeled data. To address this issue, we propose a simple and effective contrastive learning-based training strategy in which we first pretrain the network using a pixel-wise, label-based contrastive loss, and then fine-tune it using the cross-entropy loss. This approach increases intra-class compactness and inter-class separability, thereby resulting in a better pixel classifier. We demonstrate the effectiveness of the proposed training strategy using the Cityscapes and PASCAL VOC 2012 segmentation datasets.\ Our results show that pretraining with the proposed contrastive loss results in large performance gains (more than 20\% absolute improvement in some settings) when the amount of labeled data is limited. In many settings, the proposed contrastive pretraining strategy, which does not use any additional data, is able to match or outperform the widely-used ImageNet pretraining strategy that uses more than a million additional labeled images.

\end{abstract}

\section{Introduction}
In the recent past, various approaches based on Convolutional Neural Networks (CNNs)~\cite{deeplab,deeplabv3+,pspnet} have reported excellent results on several semantic segmentation datasets by first pretraining their models on the large-scale ImageNet~\cite{imagenet} classification dataset and then fine-tuning them with large amounts of pixel-level annotations. This training strategy has several disadvantages: First, collecting a large, pixel-level annotated dataset is time-consuming and expensive. For example, the average time taken to label a single image in the Cityscapes dataset is 90 minutes~\cite{cityscapes}. Second, the ImageNet dataset can only be used for non-commercial research, making the ImageNet pretraining strategy unsuitable for building real-world products. Collecting a proprietary large-scale classification dataset similar to ImageNet would be expensive and time-consuming.
Third, ImageNet pretraining does not necessarily help segmentation of non-natural images, such as medical images~\cite{raghu2019transfusion}. 

To reduce the need for large amounts of dense pixel-level annotations and the additional large-scale labeled ImageNet dataset, this work focuses on training semantic segmentation models using only a limited number of pixel-level annotated images (no ImageNet dataset). This is challenging since CNN models can easily overfit to limited training data.

\begin{figure}[t]
    \centering
    \includegraphics[trim=45 110 20 20,clip,width=0.45\textwidth]{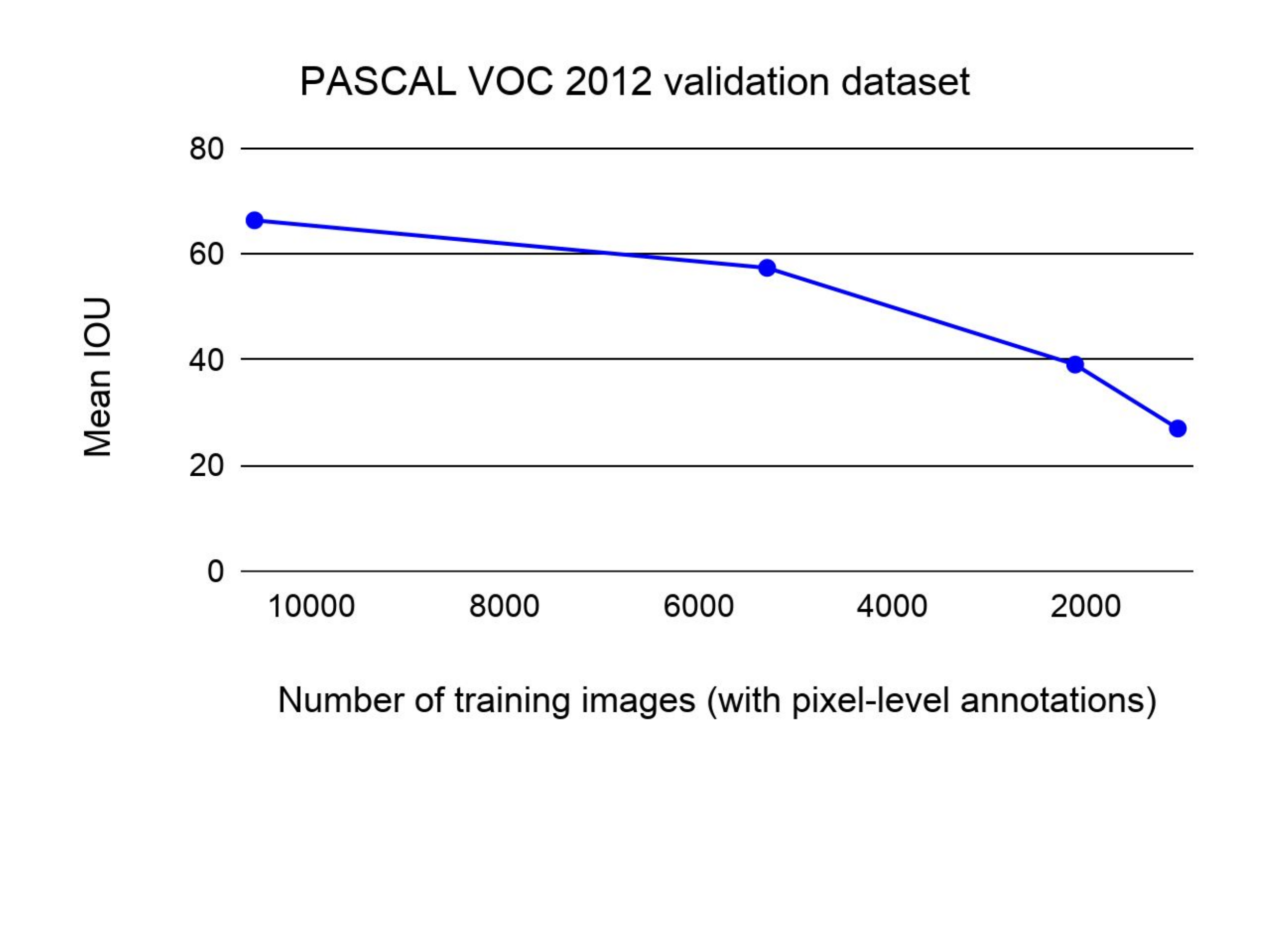}
    \caption{When trained with pixel-wise cross-entropy loss, the performance of a semantic segmentation model drops significantly as the number of labeled training images decreases.\ Here, we use a DeepLabV3+~\cite{deeplabv3+} model with the ResNet50-based encoder of~\cite{deeplabv3}.}
    \label{fig:pascal_drop}
\end{figure}

Typical semantic segmentation models consist of a deep CNN feature extractor followed by a pixel-wise softmax classifier, and are trained using a pixel-wise cross-entropy loss.\ While these models perform well when trained with a large number of pixel-level annotated images, their performance drops significantly as the number of labeled training images decreases (see Fig.~\ref{fig:pascal_drop}).
This happens because CNNs trained with the cross-entropy loss can easily overfit to small amounts of labeled data, as the cross-entropy loss focuses on creating class decision boundaries and does not explicitly encourage intra-class compactness or large margins between classes~\cite{ElsayedKMRB18,LiuWYY16,SunCWL15}.

\begin{figure}[t]
\centering
    \includegraphics[trim=60 40 50 50,clip,width=0.42\textwidth]{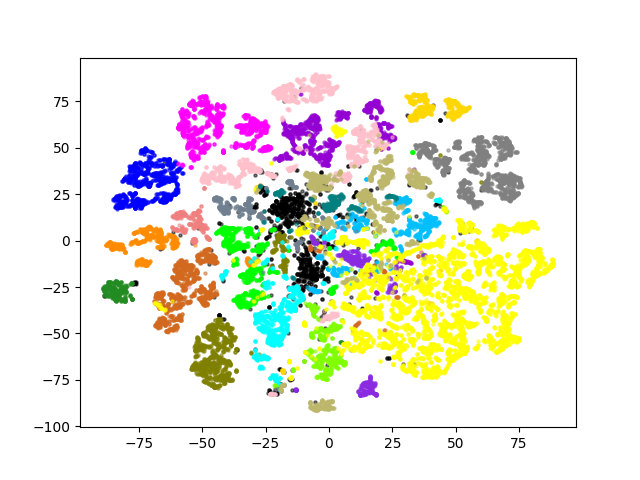}
    \caption{Distribution of various classes in the softmax input feature space of a model trained using only cross-entropy loss on 2118 labeled images from the PASCAL VOC 2012 dataset.}
    \label{fig:emb_pascal20_baseline}
\end{figure}

\begin{figure}[t]
\centering
    \includegraphics[trim=60 40 50 50,clip,width=0.42\textwidth]{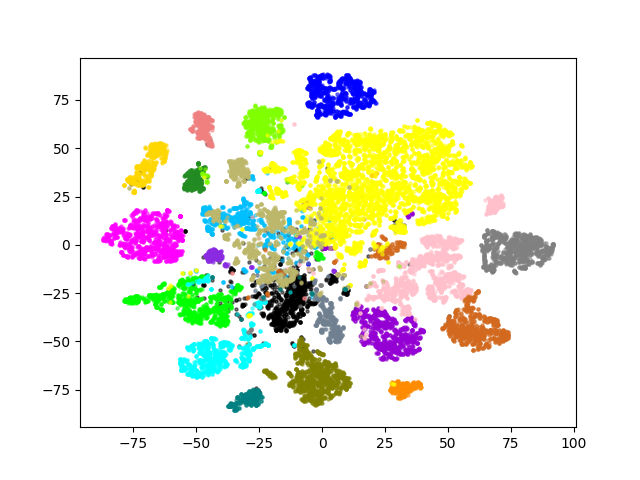}
    \caption{Distribution of various classes in the softmax input feature space of a model trained using the proposed training strategy on 2118 labeled images from the PASCAL VOC 2012 dataset.}
    \label{fig:emb_pascal20_proposed}
\end{figure}

To address this issue, we propose to first pretrain the feature extractor using a pixel-wise, label-based contrastive loss (referred to as \emph{contrastive pretraining}), and then fine-tune the entire network including the pixel-wise softmax classifier using the cross-entropy loss (referred to as \emph{softmax fine-tuning}).\ This approach increases both intra-class compactness and inter-class separability as the label-based contrastive loss~\cite{supcon} encourages the features of pixels from the same class to be close to each other and the features of pixels from different classes to be far away.\ The increased intra-class compactness and inter-class separability naturally lead to a better pixel classifier in the fine-tuning stage. Figures~\ref{fig:emb_pascal20_baseline} and~\ref{fig:emb_pascal20_proposed} show the distributions of various classes in the softmax input feature spaces of models trained with the cross-entropy loss and the proposed strategy, respectively, using 2118 labeled images from the PASCAL VOC 2012 dataset. The mean IOU values of the corresponding models on the PASCAL VOC 2012 validation dataset are 39.1 and 62.7, respectively. The class support regions are more compact and separated when trained with the proposed strategy, leading to a better performance. We use t-SNE~\cite{tsne} for generating the visualizations.

Various existing semi-supervised and weakly-supervised semantic segmentation approaches also focus on reducing the need for pixel-level annotations by leveraging additional unlabeled images~\cite{ChenLCCCZAS20,abs-1906-01916,HungTLL018,OualiHT20} or weaker forms of annotations such as bounding boxes~\cite{boxsup,Khoreva_2017_CVPR, PapandreouCMY15,Song_2019_CVPR} and image-level labels~\cite{ Araslanov_2020_CVPR,LeeKLLY19,PapandreouCMY15}. In contrast to these approaches, the proposed contrastive pretraining strategy does not use any additional data, and is complimentary to them.

Pixel-wise cross-entropy loss ignores the relationships between pixels. To address this issue, region-based loss functions such as region mutual information loss~\cite{zhao2019region} and affinity field loss~\cite{ke2018adaptive} have been proposed. Different from these loss functions which model pixel relationships in the label space, the proposed contrastive loss models pixel relationships in the feature space. Also, while these loss functions only model relationships between pixels within a local neighborhood, the proposed loss encourages the features of same class pixels to be similar and features of different class pixels to be dissimilar irrespective of their image locations.

Some recent works such as~\cite{ChaitanyaEKK20,PinheiroABGC20, abs-2011-09157, abs-2102-04803, abs-2011-10043} also used pixel-wise contrastive loss for the task of semantic segmentation. However, these works focus on leveraging unlabeled data through self-supervised contrastive learning, and make use of the labels only in the fine-tuning stage. In contrast, we focus on supervised contrastive learning, and make use of the labels in both pretraining and fine-tuning stages.

We perform experiments on two widely-used semantic segmentation benchmark datasets, namely, Cityscapes and PASCAL VOC 2012, and show that pixel-wise, label-based contrastive pretraining results in large performance gains when the amount of labeled data is limited.\\[5pt]
\noindent Our main contributions are as follows:
\begin{itemize}[topsep=2pt,itemsep=1pt]
    \item \textbf{New loss functions:} We extend supervised contrastive learning~\cite{supcon} to the task of semantic segmentation. We propose and evaluate three variants of pixel-wise, label-based contrastive loss.
    \item \textbf{Simple training approach:} We propose a simple contrastive learning-based pretraining strategy for improving the performance of semantic segmentation models. We consider the simplicity of our pretraining strategy as its main strength since it can be easily adopted by existing and future semantic segmentation approaches.
    \item \textbf{Strong results:} We show that label-based contrastive pretraining results in large performance gains on two widely-used semantic segmentation datasets when the amount of labeled data is limited.\ We also show that, in most settings, the proposed contrastive pretraining which does not use any additional data, outperforms the widely-used ImageNet pretraining which uses more than a million additional labeled images.
    \item \textbf{Detailed analyses:} We show visualizations of class distributions in the feature spaces of trained models to provide insights into why the proposed training strategy works better (Fig.~\ref{fig:emb_pascal20_baseline} and~\ref{fig:emb_pascal20_proposed}). We also present ablation studies that justify our two-stage training strategy.
\end{itemize}
\section{Related works}
\paragraph{Self-supervised contrastive learning} These approaches learn representations in a discriminative fashion by contrasting positive pairs against negative pairs. Recently, several approaches based on contrastive loss~\cite{HadsellCL06} have been proposed for self-supervised visual representation learning~\cite{simclr,simclrv2,dosovitskiy2014discriminative,He0WXG20,abs-2005-04966,WuXYL18,ye2019unsupervised}. These approaches treat each instance as a class and use contrastive loss-based instance discrimination for representation learning. Specifically, they use augmented version of an instance to form the positive pair and other randomly sampled instances to form negative pairs for the contrastive loss. Some recent works~\cite{KalantidisSPWL20,abs-2010-04592} also explored hard negative mining strategies for contrastive learning. Noting that using a large number of negatives is crucial for the success of contrastive loss-based representation learning, various recent approaches use memory banks to store the representations~\cite{He0WXG20,abs-1906-05849,WuXYL18}. Inspired by the effectiveness of self-supervised contrastive learning for image-level recognition tasks, various recent approaches extended it to pixel-level prediction tasks~\cite{ChaitanyaEKK20,PinheiroABGC20, abs-2011-09157, abs-2102-04803, abs-2011-10043}.\\\vspace{-15pt}
\paragraph{Supervised contrastive learning} Recently,~\cite{supcon} proposed supervised contrastive loss for the task of image classification.\ This loss can be seen as a generalization of the widely-used metric learning losses such as  N-pairs~\cite{npairloss} and  triplet~\cite{tripletloss} losses to the scenario of multiple positives and negatives generated using class labels. Different from~\cite{supcon}, this work focuses on much tougher pixel-level semantic segmentation task, and proposes three variants of pixel-wise, label-based contrastive loss. Since collecting labeled data for the task of semantic segmentation is difficult, we focus on the limited labeled data setting, and show that label-based contrastive learning is highly effective.\ Concurrent to this work, (still unpublished) ~\cite{abs-2101-11939} also introduced a pixel-wise, label-based contrastive loss for semantic segmentation. However,~\cite{abs-2101-11939} trains segmentation models using both cross-entropy and contrastive losses simultaneously, which is different from our contrastive pretraining followed by softmax fine-tuning strategy. Our experiments show that the proposed two-stage training is more effective than joint training. Also,~\cite{abs-2101-11939} does not demonstrate the effectiveness of contrastive learning in the limited labeled data setting.\\\vspace{-20pt}
\paragraph{Semantic segmentation} Since CNNs have been introduced to solve the semantic segmentation problem~\cite{FarabetCNL13,fcn}, several deep CNN-based approaches have been proposed that gradually improved the performance~\cite{deeplab,deeplabv3,deeplabv3+,fu2019dual,gaussiancrf,OCRseg,yuan2018ocnet,cfnet,pspnet} using large amounts of pixel-level annotations. However, collecting dense pixel-level annotations is difficult and costly. 

To address this issue, several existing works use the large-scale ImageNet classification dataset for pretraining their models, and also leverage additional weaker forms of supervision such as image-level labels~\cite{ Araslanov_2020_CVPR,HuangWWLW18,LeeKLLY19,PapandreouCMY15}, bounding boxes~\cite{boxsup,Khoreva_2017_CVPR, PapandreouCMY15,Song_2019_CVPR}, scribbles~\cite{Lin_2016_CVPR,TangDPBS18} and points~\cite{AmyECCV2016}, or unlabeled images~\cite{ChenLCCCZAS20,abs-2004-08514,abs-1908-05724,abs-2007-07936,SoulySS17,abs-2010-09713}. In contrast to these approaches, the proposed training strategy does not require any additional data or annotations.

Another relevant line of work includes approaches that use region-based loss functions~\cite{ke2018adaptive,zhao2019region} to model pixel relationships. While~\cite{ke2018adaptive} uses a pairwise affinity loss based on KL divergence between predicted class probabilities of two pixels,~\cite{zhao2019region} uses a region Mutual Information (MI) loss that maximizes the MI between predicted and groundtruth distributions of patch labels. While these losses model pixel relationships in the label space, the proposed contrastive loss models pixel relationships in the feature space.


A few existing works~\cite{abs-1708-02551,FathiWRWSGM17,HarleyDK17,KongF18a} use metric learning based on independent pairwise similarity and dissimilarity losses for the tasks of semantic and instance segmentation. However, these works only model relationships between pixels within a local image neighborhood or an object instance. Different from these works, the proposed contrastive loss models relationships between pixels irrespective of their image locations, and contrasts a similar pair with a large number of dissimilar pairs. Also, these works do not demonstrate the effectiveness of contrastive pretraining in the limited labeled data setting.

Recently,~\cite{segsort} proposed to train the feature extractor of a semantic segmentation model by maximizing the log likelihood of extracted pixel features under a mixture of vMF distributions model. During inference, they first segment the pixel features using spherical K-Means clustering, and then perform k-nearest neighbor search for each
segment to retrieve labels from segments in the training set. While this approach is shown to improve the performance when compared to the widely-used pixel-wise softmax training, it is very complicated as it uses a two-stage expectation-maximization algorithm for training. In comparison, the proposed approach is simple, and can be easily adopted by existing and future semantic segmentation approaches.



\section{Proposed approach}
\subsection{Pixel-wise label-based contrastive loss}
In this work, we extend supervised contrastive learning to pixel-level tasks such as semantic segmentation, and propose three pixel-wise, label-based contrastive losses to pretrain a semantic segmentation model.

Let $I$ denote an image and $\hat{I}$ its distorted version (e.g., color jittering). Let $y_p^I$ denote the class label of pixel $p$ in $I$, $N_{c}^I$ denote the number of pixels in $I$ with class label $c$, and $N^I$ denote the total number of pixels in $I$. Let $f^{I}_p$ be a $d$-dimensional, unit-normalized feature extracted from $I$ at pixel $p$,  Let $\mathbbm{1}_{pk}^{AB} = \mathbbm{1}\left[y_p^A = y_k^B \right]$ and $e_{pk}^{AB} = exp \left(f_p^A \cdot f_k^B\right / \tau)$, where $\tau$ is a temperature parameter. \\\vspace{-18pt}

\paragraph{Within-image loss} Our within-image, pixel-wise, label-based contrastive loss which encourages features of pixels in an image to cluster according to their labels is given by 
\begin{equation}
    - \frac{1}{N^I}\sum_{p=1}^{N^I} \frac{1}{{N_{y_p^I}^{\hat{I}}}}\sum_{q = 1}^{N^{\hat{I}}} \mathbbm{1}_{pq}^{I\hat{I}}\  \textbf{log}\left(\frac{e_{pq}^{I\hat{I}}}{\sum\limits_{k=1}^{N^{\hat{I}}} e_{pk}^{I\hat{I}}}\right),
\label{eqn:within-loss}
\end{equation}

In our experiments, image $\hat{I}$ is generated from $I$ by applying distortions with probability $p = 0.8$. Hence, for some samples in a minibatch the contrastive loss is between features of original and distorted pixels, and for other samples, the loss is between features of original pixels. We compute this contrastive loss separately for each image, and then average it across the images in a minibatch.\\\vspace{-18pt}

\paragraph{Cross-image loss} Our cross-image, pixel-wise, label-based contrastive loss extends the within-image loss~\eqref{eqn:within-loss} by using additional positives from another image $J$. Positive pixels from $J$ can be interpreted as harder positives since they come from a different image. We do not use additional negatives from $J$ since negatives from a different image can be interpreted as easier negatives~\footnote{When we experimented with adding negatives from another image $J$, we observed some performance drop in our initial experiments.}. The cross-image loss for an image pair $I$ and $J$ is given by
\begin{multline}
    - \frac{1}{N^I}\sum_{p=1}^{N^I} \sum_{q = 1}^{N^{\hat{I}}} \frac{\mathbbm{1}_{pq}^{I\hat{I}}}{N_{y_p^I}^{\hat{I}} + N^{\hat{J}}_{y_p^I}}\  \textbf{log}\left(\frac{e_{pq}^{I\hat{I}}}{\sum\limits_{k=1}^{N^{\hat{I}}} e_{pk}^{I\hat{I}} + \sum\limits_{k=1}^{N^{\hat{J}}} \mathbbm{1}_{pk}^{I\hat{J}} e_{pk}^{I\hat{J}}}\right)\\
    - \frac{1}{N^I}\sum_{p=1}^{N^I} \sum_{q = 1}^{N^{\hat{J}}} \frac{\mathbbm{1}_{pq}^{I\hat{J}}}{N_{y_p^I}^{\hat{I}} + N^{\hat{J}}_{y_p^I}}\  \textbf{log}\left(\frac{e_{pq}^{I\hat{J}}}{\sum\limits_{k=1}^{N^{\hat{I}}} e_{pk}^{I\hat{I}} + \sum\limits_{k=1}^{N^{\hat{J}}} \mathbbm{1}_{pk}^{I\hat{J}} e_{pk}^{I\hat{J}}}\right).
\label{eqn:across-loss}
\end{multline}

For each image in a minibatch, we compute the cross-image contrastive loss by pairing the image with another random image from the minibatch, and average the loss across all the images.\\\vspace{-18pt}
\paragraph{Batch loss} We also considered a batch variant that treats all the pixels in a minibatch as a single bag of pixels for computing the contrastive loss~\eqref{eqn:within-loss}.\ While one would expect this batch variant to outperform within-image and cross-image variants (due to interactions across multiple images), our experimental results indicate the opposite. Please see Section~\ref{sec:performance_gain} for further details.

\subsection{Proposed training strategy}
\begin{figure}[t]
    \centering
    \includegraphics[trim=20 230 60 20,clip,width=0.45\textwidth]{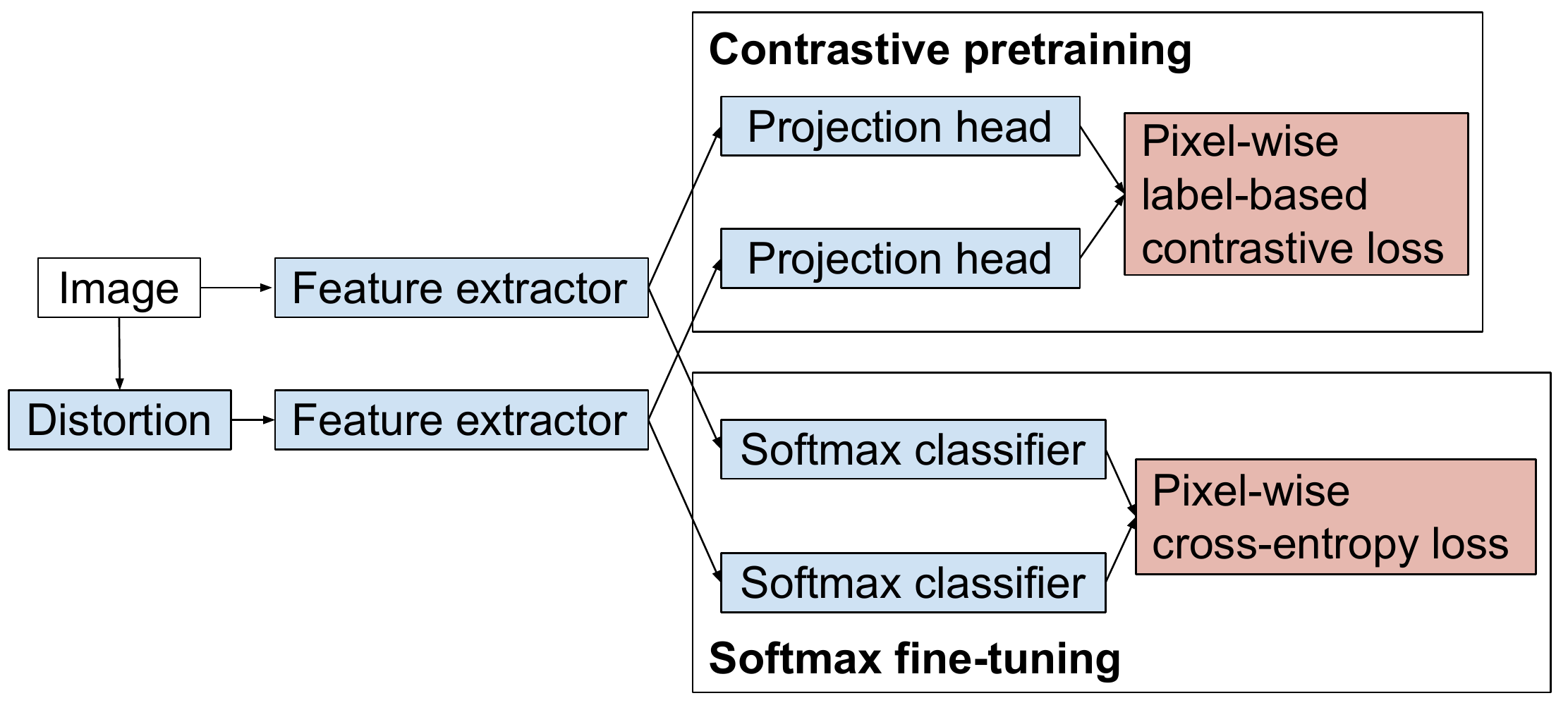}
    \caption{Proposed training strategy: Pixel-wise supervised contrastive pretraining followed by softmax fine-tuning.}
    \label{fig:approach}
\end{figure}

Typical semantic segmentation models consist of a deep CNN feature extractor followed by a pixel-wise softmax classifier. We first pretrain the CNN feature extractor from scratch with a pixel-wise, class label-based contrastive loss. Following~\cite{simclr,simclrv2,supcon} we use a projection head while training with contrastive loss, i.e., the features $f_p^I$ used in loss~\eqref{eqn:within-loss} and~\eqref{eqn:across-loss} are the outputs of a projection head rather than the original feature extractor (see Fig.~\ref{fig:approach}). After contrastive pretraining, we discard the projection head, add a pixel-wise softmax classifier on top of the feature extractor, and fine-tune the entire network with pixel-wise cross-entropy loss. 

Note that there is no interaction between pixels from different images in the within-image contrastive loss. So, it is crucial to train the entire network in the fine-tuning stage. While within-image loss-based contrastive pretraining encourages pixels within an image to cluster according to their labels, softmax fine-tuning rearranges these clusters so that they fall on the correct side of the decision boundary. 

\section{Experiments}
\subsection{Datasets and metrics}
\noindent \textbf{PASCAL VOC 2012}~\cite{pascalvoc}: This dataset consists of 10,582 training, 1,449 validation, and 456 test images with annotations for one background  and 20 foreground object classes. The performance is measured in terms of pixel Intersection-Over-Union (IOU) averaged across the 21 classes.\\[5pt]
\noindent \textbf{Cityscapes}~\cite{cityscapes}: This dataset consists of 2975 training, 500 validation, and 1525 test images. Following the evaluation protocol of~\cite{deeplab}, 19 semantic labels are used for evaluation, and the void label is ignored. The performance is measured in terms of pixel IOU averaged across the 19 classes. 

All the results reported in this section correspond to the validation splits of these datasets. Please refer to the supplementary material for results on the test splits.


\subsection{Model architecture}
Our feature extractor follows DeepLabV3$+$~\cite{deeplabv3+} encoder-decoder architecture with the ResNet50-based encoder of DeepLabV3~\cite{deeplabv3}. The output spatial resolution of the feature extractor is four times lower than the input resolution.
 Our projection head consists of three $1\times 1$ convolution layers with 256 channels followed by a unit-normalization layer. The first two layers in the projection head use the ReLU activation function. 
  \begin{figure}[t]
    \centering
    \includegraphics[trim=30 100 40 30,clip,width=0.48\textwidth]{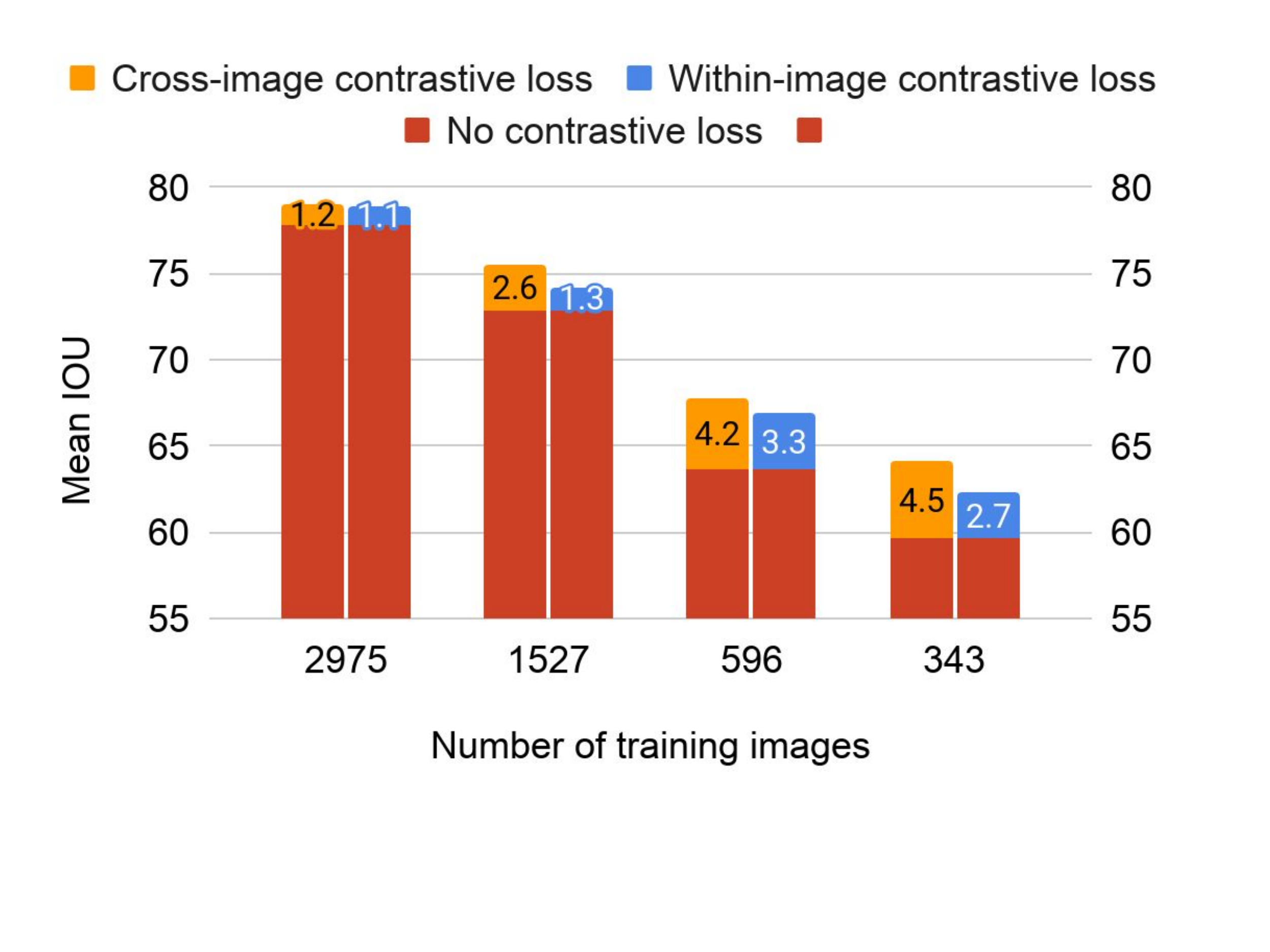}
    \caption{Improvement on the Cityscapes dataset due to contrastive pretraining.}
    \label{fig:cityscapes_fs}
\end{figure}
 \subsection{Training and inference}
 Following~\cite{deeplabv3,deeplabv3+}, we use $513\times 513$ random crops extracted from preprocessed (random left-right flipping and scaling) input images for training. All the models are trained from scratch using stochastic gradient descent on 8 replicas with minibatches of size 16, momentum of 0.9, weight decay of $4e^{-5}$,  and cosine learning rate decay. When we use softmax training without contrastive pretraining, we use an initial learning rate of 0.03 and 600K~\footnote{600K steps were chosen after trying 300K, 600K and 1M with different learning rates.} training steps when the 
number of labeled images is above 2500 in the case of PASCAL VOC 2012 dataset and above 1000 in the case of Cityscapes dataset, and 300K training steps in other settings\textsuperscript{\ref{note1}}.
 For contrastive pretraining, we use an initial learning rate of 0.1 and 300K training steps. For softmax fine-tuning after contrastive pretraining, we use an initial learning rate of 0.007 and 300K training steps except when the number of labeled images is below 2500 in the case of PASCAL VOC 2012. In this case, we use 50K training steps\footnote{\label{note1} We observed overfitting with longer training when the number of labeled images is low.}.  The temperature $\tau$ of contrastive loss is set to 0.07.  We use color distortions from~\cite{simclr} for contrastive pretraining, and random brightness and contrast adjustments for softmax fine-tuning~\footnote{Using hue and saturation adjustments from~\cite{simclr} while training the softmax classifier resulted in a slight drop in the performance.}. 

For $513\times 513$ input, our feature extractor produces a $129\times 129$ feature map. Since the memory complexity of our contrastive loss is quadratic in the number of pixels, to avoid GPU memory issues, we resize the feature map to $65\times 65$ using bilinear resizing before computing the contrastive loss. The corresponding low-resolution label map is obtained from the original label map using nearest neighbor downsampling. For softmax training, we follow~\cite{deeplabv3+} and upsample the logits from $129\times 129$ to $513\times 513$ using bilinear resizing before computing the pixel-wise cross entropy loss.

Since the model is fully-convolutional, during inference, we directly run it on an input image and upsample the output logits to input resolution using bilinear resizing.

\begin{figure}[t]
    \centering
    \includegraphics[trim=20 0 20 20,clip,width=0.48\textwidth]{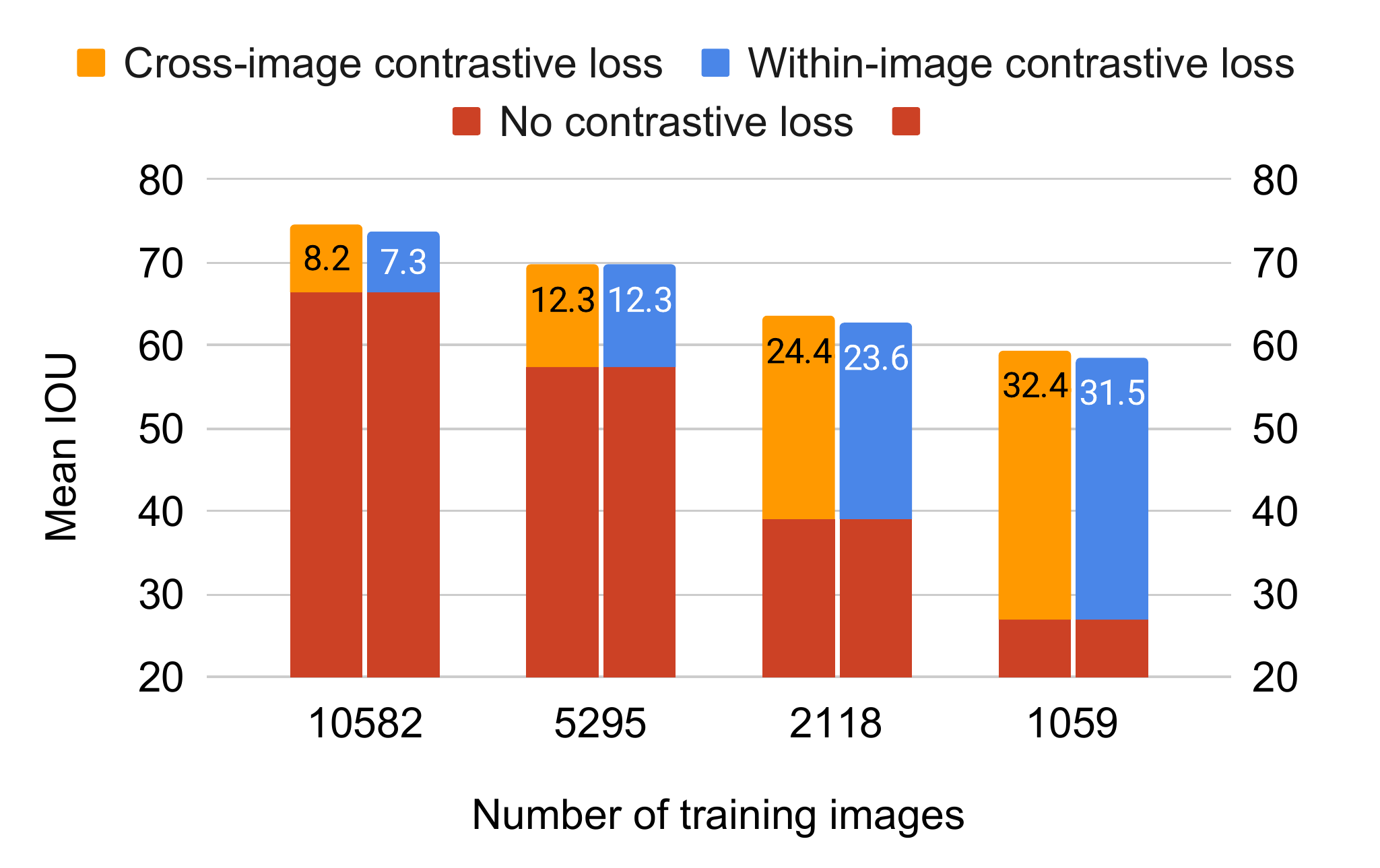}
    \caption{Improvement on the PASCAL VOC 2012 dataset due to contrastive pretraining.}
    \label{fig:pascal_fs}
\end{figure}

\subsection{Performance gain by contrastive pretraining}
\label{sec:performance_gain}
Figures~\ref{fig:cityscapes_fs} and~\ref{fig:pascal_fs} show the performance improvements on the validation splits of Cityscapes and PASCAL VOC 2012 datasets, respectively, obtained by contrastive pretraining. Both within-image and cross-image contrastive loss-based pretraining consistently improve the performance on both the datasets for different amounts of training data. On the Cityscapes dataset, we see large gains (more than 4 points) when the number of labeled training images is less than 600, and a decent gain (1.2 points) even when using the entire training set of 2975 images. On the PASCAL VOC 2012 dataset, we see huge gains (up to about 30 points) for all label counts, and we are able to reduce the labeling requirements by $2\times$ while improving the performance.\ Notably, by using 1059 images, we are able to outperform the model trained using only cross-entropy loss on $5\times$ more data (5295 images). These results clearly demonstrate the effectiveness of the proposed contrastive pretraining. 

Cross-image contrastive loss outperforms within-image contrastive loss since it makes use of positives from other images which can be seen as harder positives when compared to within-image positives. In most of the settings, the cross-image loss outperforms the within-image loss by 0.8 or more points. Specifically, on the Cityscapes dataset, the cross-image loss outperforms the within-image loss by 1.8 points when only 343 labeled images are available. 

We also conducted experiments with the batch variant of our contrastive loss which considers all the pixels in a minibatch as a single bag of pixels for computing the loss~\eqref{eqn:within-loss}. Note that the memory complexity of loss~\eqref{eqn:within-loss} is quadratic in the number of pixels. Hence, to avoid GPU memory issues, we randomly sample 10K pixels from the entire minibatch for computing the batch contrastive loss. Table~\ref{tab:loss_variants} compares the performance of the batch variant with the other two variants. While one would expect the batch variant to perform better because of the pixel interactions across multiple images, the results indicate the opposite. Note that, while we have contrastive loss terms for every pixel in the within-image and cross-image variants, only a subset of pixels contribute to the loss in the batch variant. We believe this to be the reason for the poor performance of the batch variant. In the near future, we plan to explore hybrid variants that will have loss terms for as many pixels as possible while still forming pixel pairs across multiple images.
\begin{figure}[t]
    \centering
    \includegraphics[trim=20 0 20 20,clip,width=0.48\textwidth]{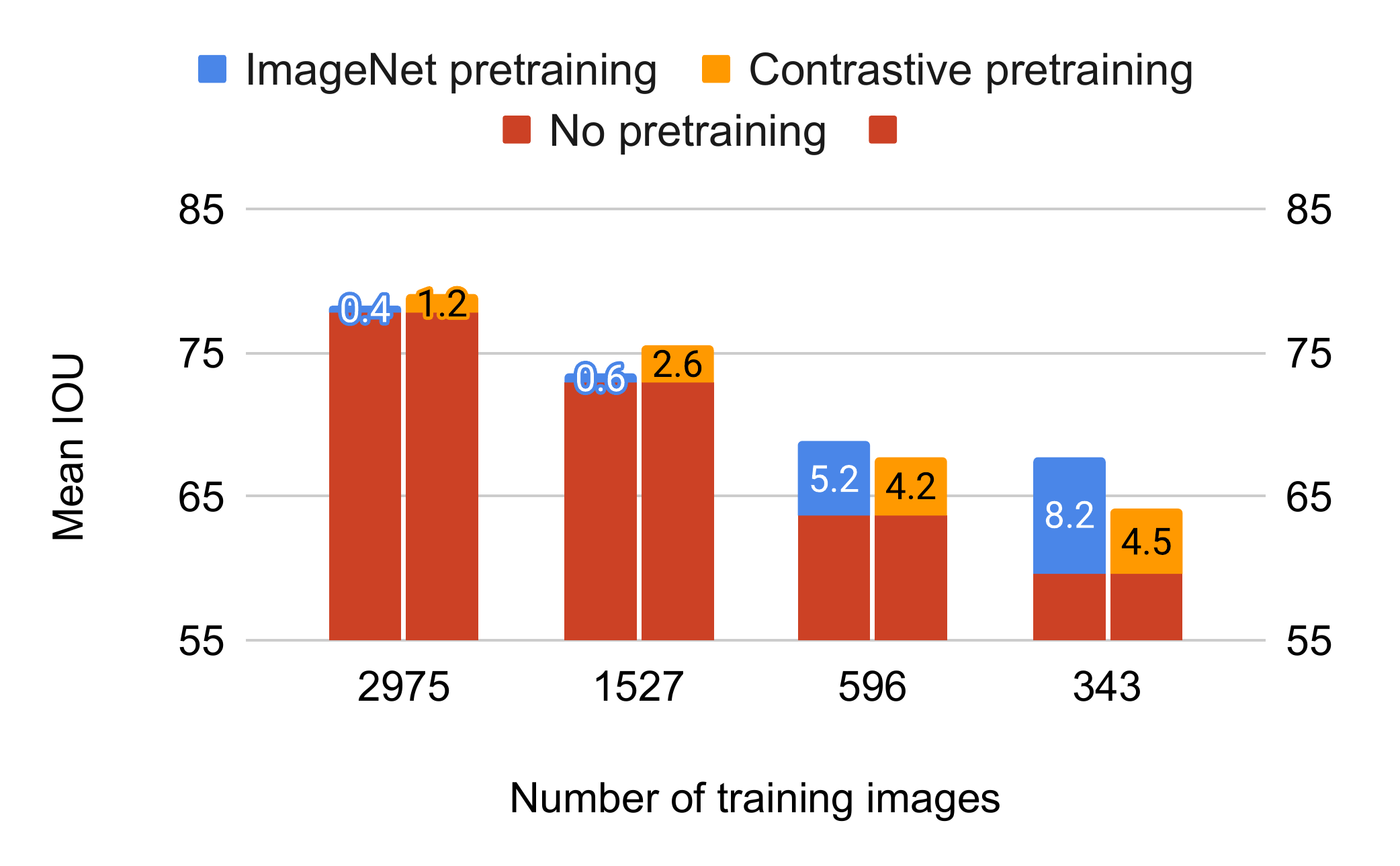}
    \caption{Improvement on the Cityscapes dataset due to different pretraining strategies.}
    \label{fig:cityscapes_imagenet}
\end{figure}

The performance improvements seen on the PASCAL VOC 2012 dataset are much higher than those seen on the Cityscapes dataset.\ We conjecture that this is because of the presence of an additional background category in the PASCAL VOC 2012 dataset. This category is comprised of diverse visual content from a wide variety of object classes (everything other than the 20 foreground object classes). Hence, explicit enforcement of intra-class compactness and inter-class separability by contrastive pretraining is helping more in the case of PASCAL VOC 2012 dataset. To verify our conjecture experimentally, we trained the segmentation model on PASCAL VOC 2012 dataset (2118 labeled images) ignoring the background category. When evaluated on the foreground categories, within-image loss-based contrastive pretraining improved the mean IOU by 3.6 points, which is much lower than the 23.6 points gain achieved in the presence of background category. This suggests that the presence of additional background category is contributing to the huge gains on the PASCAL VOC 2012 dataset.
\begin{figure}[t]
    \centering
    \includegraphics[trim=20 0 20 20,clip,width=0.48\textwidth]{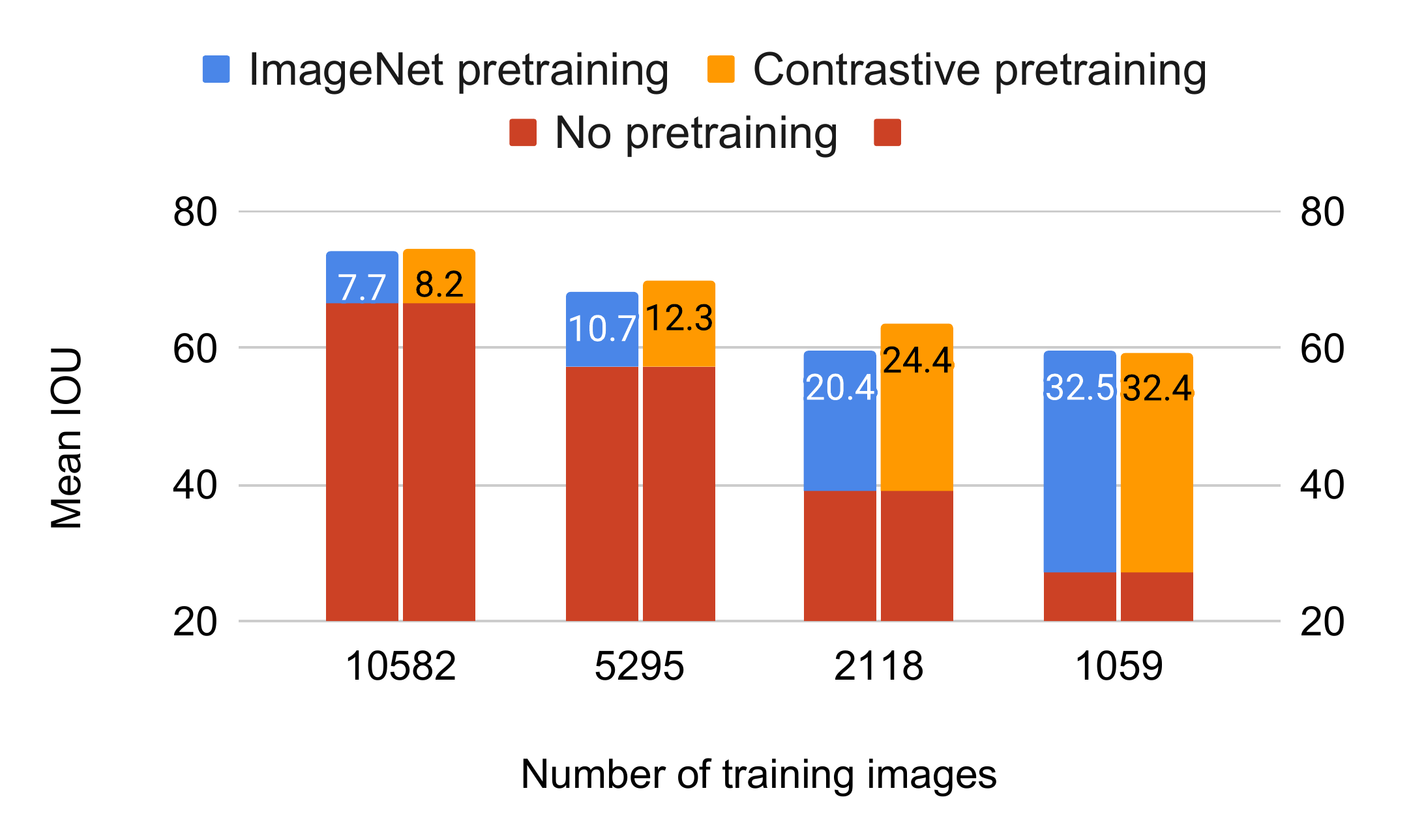}
    \caption{Improvement on the PASCAL VOC 2012 dataset due to different pretraining strategies.}
    \label{fig:pascal_imagenet}
\end{figure}
\begin{table}[t]
\renewcommand{\arraystretch}{1.2}
\centering
\small
\caption{Performance of different contrastive loss variants.}
\label{tab:loss_variants}
\begin{tabular}{lrr}
\toprule
    Dataset & Cityscapes & PASCAL VOC\\
    Training images & 596 images & 2118 images\\
    \midrule
    No contrastive loss & 63.6 & 39.1\\\midrule
    Batch loss & ($\uparrow$ 2.4) 66.0 & ($\uparrow$ 22.6) 61.7 \\
    Within-image loss & ($\uparrow$ 3.3) 66.9 & ($\uparrow$ 23.6) 62.7 \\
    Cross-image loss & ($\uparrow$ \textbf{4.2}) \textbf{67.8} & ($\uparrow$ \textbf{24.4}) \textbf{63.5}\\
    \bottomrule
\end{tabular}
\end{table}
\subsection{Comparison with ImageNet pretraining}
\label{sec:imagenet-comparison}
Most of the existing semantic segmentation approaches pretrain their models on the large-scale ImageNet classification dataset~\cite{imagenet} to achieve state-of-the-art results. Even works such as~\cite{HeGD19,ZophGLCLC020} which show that ImageNet pretraining can be omitted for the task of object detection on some datasets, acknowledge that ImageNet pretraining is important for semantic segmentation.\ While it can lead to significant performance gains, ImageNet pretraining may not be used for building commercial products.\ Collecting such a large-scale proprietary dataset is also time-consuming and expensive. In contrast, the proposed strategy achieves performance gains without using any additional data. Figures~\ref{fig:cityscapes_imagenet} and~\ref{fig:pascal_imagenet} compare the performances of ImageNet-pretrained and contrastive-pretrained models. Contrastive pretraining matches or outperforms ImageNet pretraining in most of the cases except when the number of labeled images is below 600 for the Cityscapes dataset. This is a significant result given that the proposed contrastive pretraining does not use any additional data and ImageNet pretraining uses more than a million additional labeled images.

\subsection{Ablation studies}
\label{sec:ablation_studies}
In this section, we perform ablation studies on the Cityscapes (596 training images) and PASCAL VOC 2012 (2118 training images) datasets using the within-image contrastive loss.
\subsubsection{Joint training}
The proposed approach first pretrains the feature extractor using a label-based contrastive loss, and then fine-tunes the entire network using the cross-entropy loss. An alternative training strategy is to train with both losses at the same time. Table.~\ref{tab:joint-training} compares these two training strategies. While joint training with both losses performs slightly better than training with only cross-entropy loss, it performs significantly worse when compared to the proposed approach~\footnote{Contrastive loss weight for joint training was chosen using grid search.}.

\begin{table}[t]
\renewcommand{\arraystretch}{1.2}
\centering
\small
\caption{Comparison of different training strategies.}
\label{tab:joint-training}
\begin{tabular}{crr}
\toprule
    \multirow{2}{*}{Strategy} & Cityscapes & PASCAL VOC\\
    & (596 images) & (2118 images)\\
    \midrule
    Only cross-entropy & 63.6 & 39.1\\
    \midrule
    Joint training & ($\uparrow$ 1.0) 64.6 & ($\uparrow$ 0.7) 39.8\\
    Proposed & ($\uparrow$ \textbf{3.3}) \textbf{66.9} & ($\uparrow$ \textbf{23.6}) \textbf{62.7}\\
    \bottomrule
\end{tabular}
\end{table}
\begin{table}[t]
\renewcommand{\arraystretch}{1.2}
\centering
\small
\caption{Effect of distortions for contrastive pretraining.}
\label{tab:distortions}
\begin{tabular}{crr}
\toprule
    \multirow{2}{*}{Distortions} & Cityscapes & PASCAL VOC\\
    & (596 images) & (2118 images)\\
    \midrule
    \xmark & 66.5 & 62.9\\
    \cmark & (1.0 $\uparrow$) 67.5 & 62.7\\
    \bottomrule
\end{tabular}
\end{table}
\subsubsection{Importance of distortions for contrastive loss}
In the case of contrastive loss-based self-supervised learning, distortions are necessary to generate positive pairs~\cite{simclr}. But, in the case of label-based contrastive learning, positive pairs can be generated using labels, and hence, it is unclear how important distortions are. In this work, we use the color distortions from a recent self-supervised learning method~\cite{simclr} that worked well for the downstream task of image classification.\  Table~\ref{tab:distortions} shows the effect of using these distortions in the contrastive pretraining stage. We can see a small performance gain on the Cityscapes dataset and no gain on the PASCAL VOC 2012 dataset~\footnote{Differences lower than 0.5 are too small to draw any conclusion.}. These results suggest that distortions that work well for image recognition may not work for semantic segmentation.\ This also warrants a careful study of various distortions to find the ones that are most suitable for the task of semantic segmentation.

\subsection{Additional results}
\subsubsection{Contrastive pretraining with OCR~\cite{OCRseg} approach}
While we use DeepLabV3+~\cite{deeplabv3+} as the baseline model in all our experiments, the proposed contrastive pretraining strategy can be easily adopted by other existing or future semantic segmentation models. To demonstrate this, we combine (within-image) contrastive loss-based pretraining strategy with the recent OCR~\cite{OCRseg} approach using the code provided by its authors~\footnote{\url{https://github.com/openseg-group/openseg.pytorch?v=2}}. Table~\ref{tab:ocrseg-results} shows the corresponding results on the Cityscapes dataset. Contrastive pretraining leads to significant performance gains demonstrating that it can be effective with multiple segmentation models.
\begin{table}[t]
\renewcommand{\arraystretch}{1.2}
\centering
\small
\caption{Contrastive pretraining with OCR~\cite{OCRseg} approach.}
\label{tab:ocrseg-results}
\begin{tabular}{crr}
\toprule
    \multirow{2}{*}{Contrastive pretraining} & \multicolumn{2}{c}{Cityscapes}\\\cline{2-3}
    & (343 images) & (596 images)\\
    \midrule
    \xmark & 57.1 & 62.2\\
    \cmark & ($\uparrow$ 6.3) 63.4 & ($\uparrow$ 3.4) 65.6 \\
    \bottomrule
\end{tabular}
\end{table}

Note that the OCR~\cite{OCRseg} approach performs worse than our DeepLab V3+ baseline (62.2 vs 63.6 for 596 training images and 57.1 vs 59.6 for 343 training images). This may be because the OCR model has more learnable parameters, and is more prone to overfitting when the number of training images is low.

\subsubsection{Comparison with region-based loss functions}
As pixel-wise cross-entropy loss ignores the relationships between pixels, region-based loss functions~\cite{ke2018adaptive,zhao2019region} have been proposed which model pixel relationships in the label space. Different from these loss functions, the proposed contrastive loss models pixel relationships in the feature space. Table~\ref{tab:region-losses} compares the proposed approach with~\cite{ke2018adaptive,zhao2019region}. For fair comparison, we train~\cite{ke2018adaptive,zhao2019region} with ResNet50-based DeepLabV3+ model using the code provided by authors of~\cite{zhao2019region}~\footnote{\url{https://github.com/ZJULearning/RMI}}. The proposed training approach (which does not use any additional data) clearly outperforms the region-based loss functions~\cite{ke2018adaptive,zhao2019region} even when they use ImageNet pretraining. This suggests that modeling pixel relationships with a loss in the feature space is more effective than modeling with losses in the label space. 

\begin{table}[t]
\renewcommand{\arraystretch}{1.2}
\centering
\small
\caption{Comparison with region loss-based approaches. Here, IN and CT refer to ImageNet and contrastive pretraining, respectively.}
\label{tab:region-losses}
\begin{tabular}{cccrr}
\toprule
    \multirow{2}{*}{Approach} & \multicolumn{2}{c}{Pretraining} & \multicolumn{2}{c}{PASCAL VOC}\\\cline{2-3}\cline{4-5}
    & IN & CT & (1059 images) & (2118 images)\\
    \midrule
    \multirow{2}{*}{AF~\cite{ke2018adaptive}} & \xmark & \xmark & 27.8 & 43.0\\
    & \cmark & \xmark & 57.4 & 60.4\\
    \multirow{2}{*}{RMI~\cite{zhao2019region}} & \xmark & \xmark & 27.9 & 37.5\\
    & \cmark & \xmark & 58.0 & 61.9\\
    Proposed & \xmark & \cmark & 59.4 & 63.5 \\
    \bottomrule
\end{tabular}
\end{table}
\subsubsection{Comparison with self-supervised learning}
Recently,~\cite{abs-2011-09157, abs-2102-04803, abs-2011-10043} explored pixel-wise self-supervised contrastive learning for the task of semantic segmentation. They first pretrain their networks on the ImageNet dataset using pixel-wise self-supervised contrastive loss, and then fine-tune them end-to-end on the target semantic segmentation dataset. Using the full labeled dataset,~\cite{abs-2011-09157} reported a mean IOU of 69.4 for the PASCAL VOC 2012 dataset, and ~\cite{abs-2011-09157},~\cite{abs-2102-04803} and~\cite{abs-2011-10043} reported mean IOUs of 75.7, 76.5 and 77.2, respectively, for the Cityscapes dataset. In comparison, using the full labeled dataset, we achieve 79.0 and 74.6 on the Cityscapes and PASCAL VOC 2012 datasets, respectively~\footnote{This is not necessarily a fair comparison since the network architectures used by~\cite{abs-2011-09157, abs-2102-04803, abs-2011-10043} are different from ours, and these approaches use additional unlabeled ImageNet dataset.}. In fact, we match the results of~\cite{abs-2011-09157} by using 50\% fewer labeled images (69.7 for PASCAL using 5.9K images and 75.5 for Cityscapes using 1.5K images). 
\subsubsection{Semi-supervised setting}
While we mainly focus on the supervised setting, the proposed training strategy can be easily extended to the semi-supervised setting where we have access to additional unlabeled images. We demonstrate this using a simple pseudo labeling strategy. In this setting, we first train a model with labeled images using the proposed approach. Then, we generate pseudo labels for the unlabeled images by running the trained model on them and thresholding the output predictions. Specifically, we assign a pixel to a class if that class receives the highest score for that pixel and that score is above a threshold T~\footnote{For our experiments, we use a threshold of 0.8 for all the foreground classes of the PASCAL VOC 2012 and Cityscapes datasets, and a threshold of 0.97 for the background class of the PASCAL VOC 2012 dataset.}. If the model does not produce a score above T for any of the classes for a pixel, then that pixel is ignored. Once we generate pseudo labels for the unlabeled images, we retrain the model on both labeled and pseudo-labeled images using the proposed approach.

Table~\ref{tab:semi-supervised} compares the proposed approach with the recent semi-supervised CCT~\cite{OualiHT20} approach which has been shown to outperform several existing semi-supervised and weakly-supervised approaches. The proposed approach with contrastive pretraining outperforms CCT by a huge margin (25-35 points). When we train the proposed pseudo label-based semi-supervised approach only with cross-entropy loss, i.e., no contrastive pretraining, its performance is similar to CCT verifying that the gap between CCT and the proposed approach is mainly due to contrastive pretraining despite some architectural differences between the networks~\footnote{While CCT uses PSPNet~\cite{pspnet}, we use DeeplabV3+ model~\cite{deeplabv3+}}. Also, the proposed approach performs competitively when compared to ImageNet-pretrained CCT, reaffirming the effectiveness of supervised contrastive pretraining.

Please refer to the supplementary material for additional results in the semi-supervised setting.
\begin{table}[t]
\renewcommand{\arraystretch}{1.2}
\centering
\small
\caption{Performance in the semi-supervised setting. Here, IN and CT refer to ImageNet and contrastive pretraining, respectively.}
\label{tab:semi-supervised}
\begin{tabular}{cccrr}
\toprule
    \multirow{2}{*}{Approach} & \multicolumn{2}{c}{Pretraining} & \multicolumn{2}{c}{PASCAL VOC} \\\cline{2-3}\cline{4-5}
    & IN & CT & (1059 images) & (2118 images)\\
    \midrule
    CCT~\cite{OualiHT20} & \xmark & \xmark & 27.1 & 40.3\\
    Proposed & \xmark & \xmark & 28.6 & 41.4\\
    Proposed & \xmark & \cmark & 60.4 & 65.2\\
    \midrule
    CCT~\cite{OualiHT20} & \cmark & \xmark & 62.9 & 65.1\\
    \bottomrule
\end{tabular}
\end{table}

\section{Conclusions and future work}
Deep CNN-based semantic segmentation models trained with cross-entropy loss perform poorly when trained with limited labeled data. To address this issue, we proposed a simple and effective contrastive learning-based training strategy in which we first pretrain the feature extractor of the model using a pixel-wise label-based contrastive loss and then fine-tune the entire network including the softmax classifier using the cross-entropy loss. This training approach increases both intra-class compactness and inter-class separability, thereby enabling a better pixel classifier. We performed experiments on PASCAL VOC 2012 and Cityscapes datasets, and achieved large performance gains by using contrastive pretraining. Specifically, in many settings, the proposed contrastive pretraining strategy which does not use any additional data, matches or outperforms the widely-used supervised ImageNet pretraining strategy.

In this work, we used a pseudo labeling-based approach to leverage unlabeled images.\ In the future, we plan to explore the proposed contrastive loss in conjunction with consistency-based loss functions~\cite{abs-1906-01916,abs-2007-07936,OualiHT20} which are commonly-used for semi-supervised learning.\\\vspace{-20pt}
\paragraph{Acknowledgements} We thank Yukun Zhu and Liang-Chieh Chen from Google Research for their support with the DeepLab codebase.

{\small
\bibliographystyle{ieee_fullname}
\bibliography{main}
}

\section*{Appendix} 
\section*{A. Performance on test splits}
Table~\ref{tab:city_test} shows the performance improvements on the test splits of the Cityscapes and PASCAL VOC 2012 datasets obtained by pretraining using within-image contrastive loss in the fully-supervised setting. Similar to the results on the validation splits (Sec.\ 4.4 of the main submission), we observe significant performance gains on the test splits.
\begin{table}[hb!]
\renewcommand{\arraystretch}{1.2}
\centering
\small
\caption{Performance on the test splits of Cityscapes and PASCAL VOC 2012 datasets.}
\begin{tabular}{crr}
\toprule
    \multicolumn{3}{c}{Cityscapes} \\
    \cmidrule{1-3}
    Training images & (2975 images) & (596 images)\\
    \midrule
No pretraining & 76.3 & 64.6 \\
Contrastive pretraining & (1.8 $\uparrow$) 78.1 & (3.4 $\uparrow$) 68.0\\
    \bottomrule
     \multicolumn{3}{c}{PASCAL VOC 2012}\\
     \cmidrule{1-3}
    Training images & (10528 images) & (2118 images)\\
    \midrule
No pretraining & 67.2 & 39.4  \\
Contrastive pretraining & (7.6 $\uparrow$) 74.8 & (21.7 $\uparrow$) 61.1  \\
    \bottomrule
\end{tabular}
	\label{tab:city_test}
\end{table}
\section*{B. Performance gain in semi-supervised setting}
\begin{figure}[t]
    \centering
    \includegraphics[trim=30 120 40 35,clip,width=0.47\textwidth]{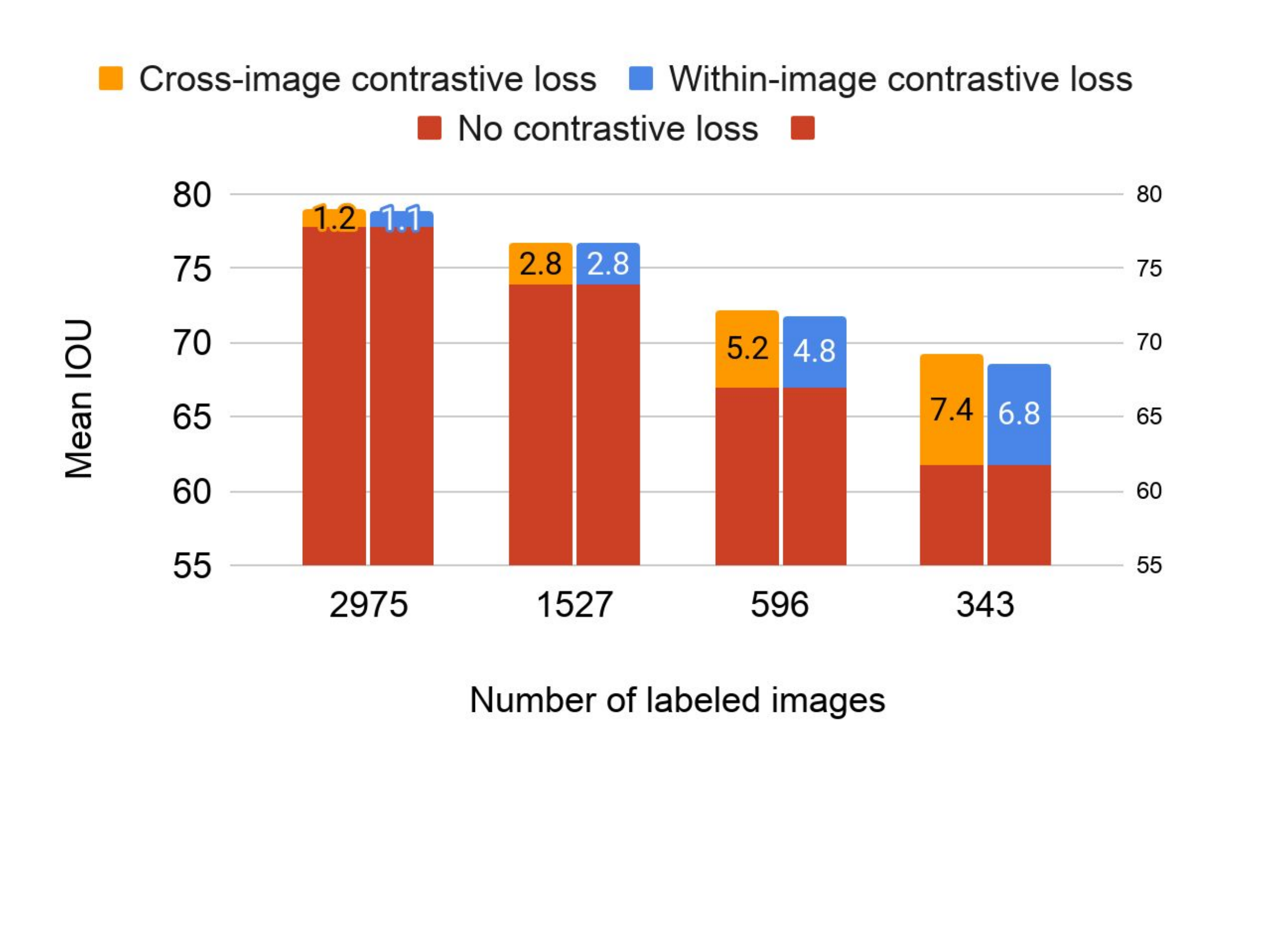}
    \caption{Improvement on Cityscapes validation dataset due to contrastive pretraining in the semi-supervised setting. Note that $\#$ unlabeled images $= 2975 - \#$ labeled images.}
    \label{fig:cityscapes_ss_supp}
    \centering
    \includegraphics[trim=30 120 40 15,clip,width=0.47\textwidth]{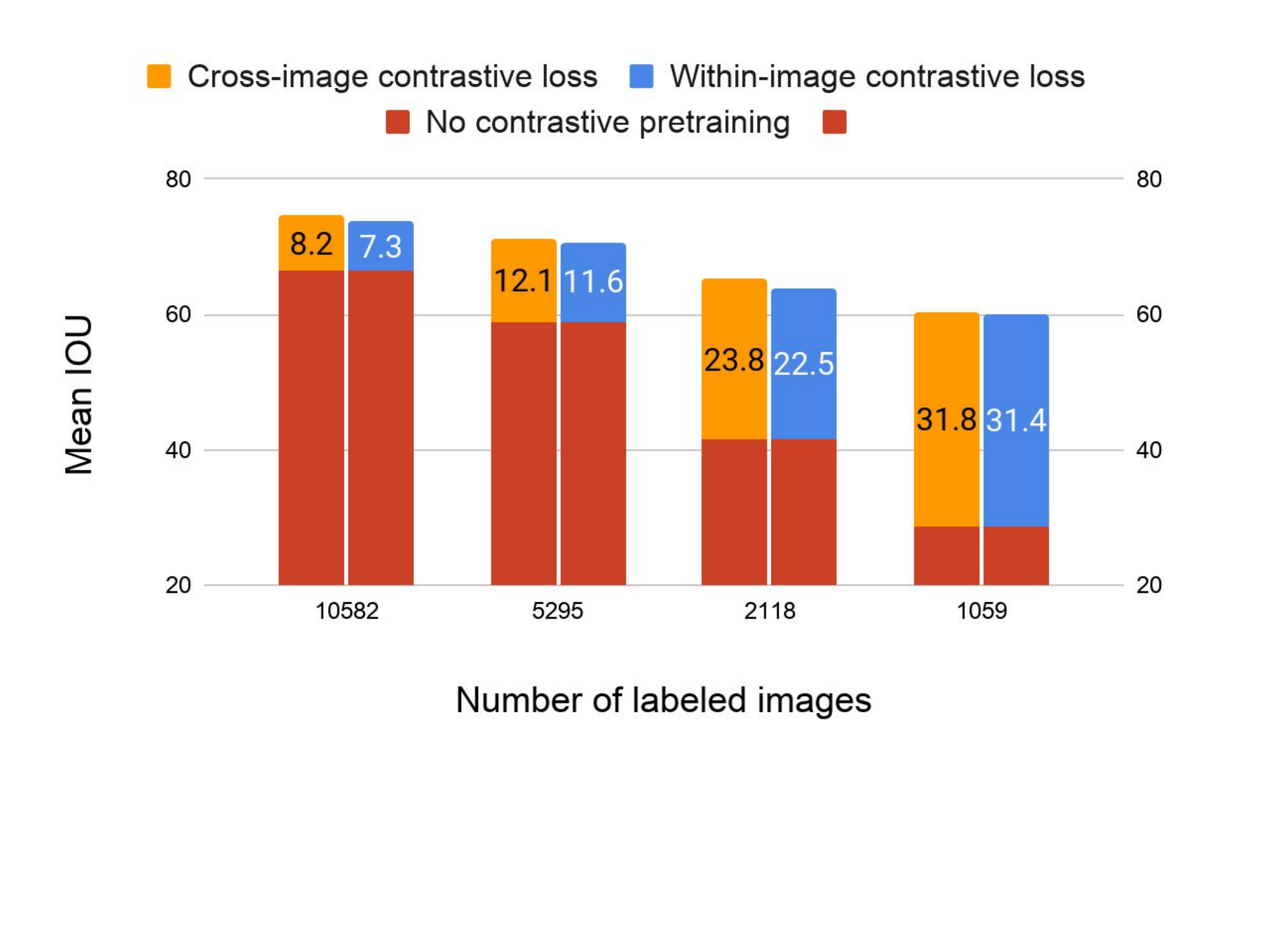}
    \caption{Improvement on PASCAL VOC 2012 validation dataset due to contrastive pretraining in the semi-supervised setting. Note that  $\#$ unlabeled images $= 10582 - \#$ labeled images.}
    \label{fig:pascal_ss_supp}
\end{figure}

Figures~\ref{fig:cityscapes_ss_supp} and~\ref{fig:pascal_ss_supp} show the performance improvements on the validation splits of the Cityscapes and PASCAL VOC 2012 datasets, respectively, obtained by contrastive pretraining in the semi-supervised setting.\ Here, we use contrastive pretraining for both the initial model that is used to generate the pseudo labels, and the final model that is trained with labeled and pseudo-labeled images. Contrastive pretraining consistently improves the performance on both the datasets for different amounts of labeled and unlabeled training data. On the Cityscapes dataset, we see large gains (2.8 - 7.4 points) in terms of mean IOU, and on the PASCAL VOC 2012 dataset, we see huge gains (up to about 30 points). Similar to the fully-supervised setting, we are able to reduce the labeling requirements by $2\times$ while improving the performance on the PASCAL VOC 2012 dataset.
\begin{figure*}[t]
    \centering
    \includegraphics[trim=80 105 80 0,clip,width=0.93\textwidth]{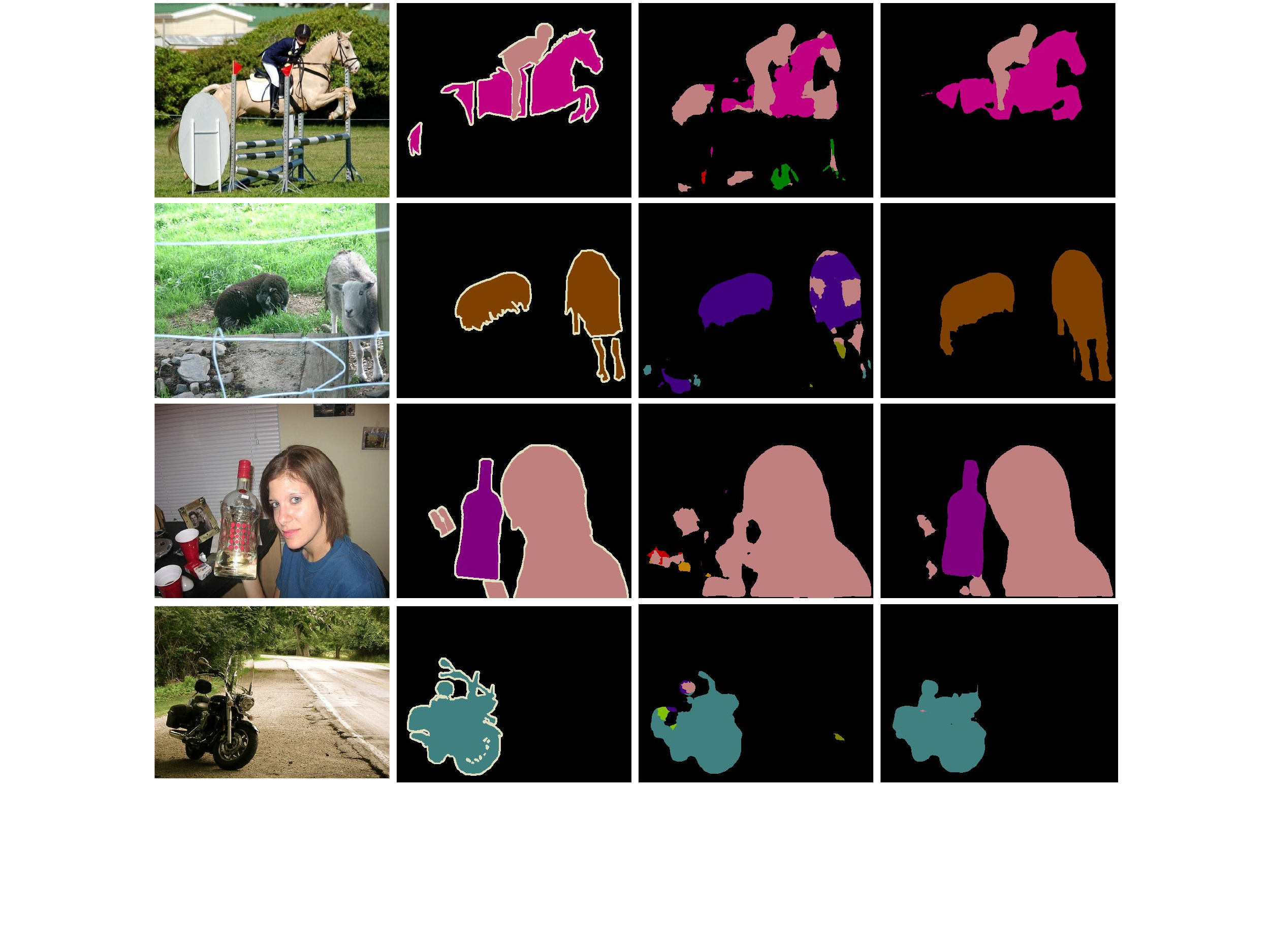}
    \includegraphics[trim=80 320 80 0,clip,width=0.93\textwidth]{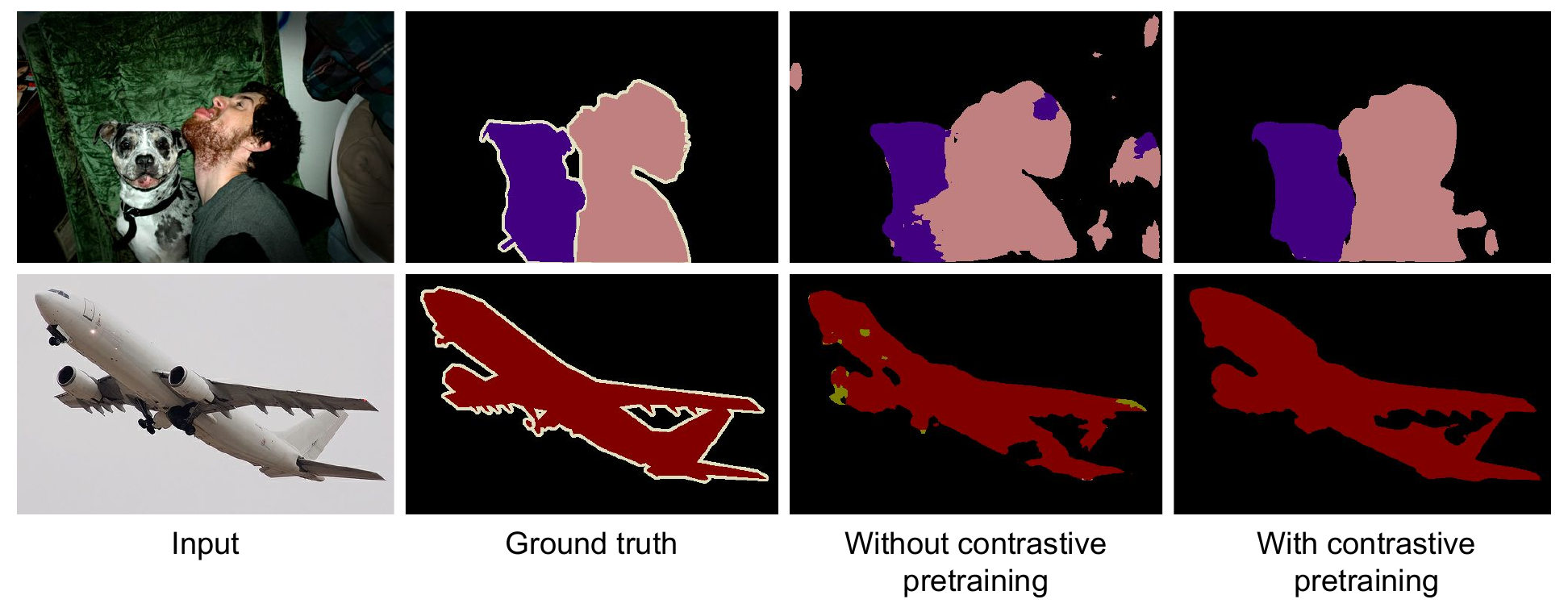}
    \caption{Comparison of models trained with and without contrastive pretraining on 2118 labeled images from PASCAL VOC 2012 dataset. Contrastive pretraining improves the results by reducing the confusion between various classes.}
    \label{fig:visual_results}    
\end{figure*}
\section*{C. Visual results}
Figure~\ref{fig:visual_results} shows some segmentation results of models trained with and without label-based contrastive pretraining using 2118 labeled images from the PASCAL VOC 2012 dataset. Contrastive pretraining improves the segmentation results by reducing the confusion between background and various foreground classes, and also the confusion between different foreground classes.

\end{document}